\documentclass[conference]{IEEEtran}
\IEEEoverridecommandlockouts
\usepackage{cite}
\usepackage{amsmath,amssymb,amsfonts}
\usepackage{algorithmic}
\usepackage{graphicx}
\usepackage{textcomp}
\usepackage{multirow}
\usepackage{xcolor}
\def\BibTeX{{\rm B\kern-.05em{\sc i\kern-.025em b}\kern-.08em
    T\kern-.1667em\lower.7ex\hbox{E}\kern-.125emX}}
\begin{document}

\title{Talent-Interview: Web-Client Cheating Detection for Online Exams\\

\thanks{Identify applicable funding agency here. If none, delete this.}
}

\author{\IEEEauthorblockN{1\textsuperscript{st} Mert Ege}
\IEEEauthorblockA{\textit{ R\&D Center} \\
\textit{Huawei Turkey}\\
Istanbul, Turkey \\
0000-0001-9060-290X}
\and
\IEEEauthorblockN{1\textsuperscript{st} Mustafa Ceyhan}
\IEEEauthorblockA{\textit{ R\&D Center} \\
\textit{Huawei Turkey}\\
Istanbul, Turkey \\
0000-0003-3268-6898}
}

\maketitle

\begin{abstract}
Online exams are more attractive after the Covid-
19 pandemic. Furthermore, during recruitment, online exams are
used. However, there are more cheating possibilities for online
exams. Assigning a proctor for each exam increases cost. At this
point, automatic proctor systems detect possible cheating status.
This article proposes an end-to-end system and submodules
to get better results for online proctoring. Object detection,
face recognition, human voice detection, and segmentation are
used in our system. Furthermore, our proposed model works
on users’ pc, meaning a client-based system. So, server cost is
eliminated. As far as we know, it is the first time the client-
based online proctoring system has been used for recruitment.
Furthermore, this cheating system works at https://www.talent-
interview.com/tr/
\end{abstract}


\section{Introduction}
Today, many jobs can be done only with a computer, regardless of the place. One of the main reasons for this is that people's work now requires information power instead of body power. There are still jobs that are done with muscle strength. However, in these professions, it turns into a machine and software. The fact that doing something individually or collectively becomes independent of the place introduces the concept of remote work. Remote working is a situation that has spread rapidly, especially during the pandemic. In this period, it has been seen that working remotely through a computer, sufficient internet infrastructure, and some applications is almost the same as going to a place or an office and even more productive in some cases \cite{remoteWorking}. 

In the traditional system, people must spend time and money going to a physical workplace. Workplaces must provide space for their employees. Employees increase the current traffic to go to their workplaces. With the widespread use of remote work, these situations can be minimized. Of course, the concept of remote work brings with it the remote work processes. At this point, it started to gain popularity in systems that make online interviews and exams. Using such systems in the recruitment processes of candidates enables candidates to enter job interviews from all over the world. Again, a similar situation is a need for educational institutions or universities to conduct their exams remotely in some cases, such as pandemics.

Online exam and recruitment systems are helpful. However, this system has a problem. The problem is that people who take the exam or interview will likely cheat. To prevent this situation, it is necessary to provide proctoring systems based on humans or software, sometimes used together. Researchers have examined such systems in \cite{hussein2020proctoring}\cite{mohammed2022proctoring}\cite{arno2021proctoring} articles.

Plagiarism control of the candidates' answers can be done with studies such as \cite{anielak2015plagiarizm} \cite{vamsi2019plagiarizm}. Nevertheless, they need to analyze the behavior of the candidate. Today's proctoring systems prefer artificial intelligence technologies for behavior analysis \cite{nigam2021ai} \cite{motwani2021ai}\cite{ataum2017ai} \cite{raj2015ai}. Artificial intelligence systems can detect some of the cheating statuses, including:
\begin{itemize}
    \item Devices that may cause cheating, 
    \item Detect background audio, 
    \item The number of candidates in front of the camera, 
    \item Detect whether the person in front of the camera is a candidate with face recognition
    \item Check whether the person has left the exam with face tracking. 
\end{itemize}

In studies such as \cite{Winarno2017AnticheatingPS}, more powerful proctoring was attempted using additional tools. However, studies that require additional tools need to be more scalable. In other studies \cite{ozgen2021ti}\cite{ozturk2021ti}, artificial intelligence tools such as face recognition, face tracking, and object detection were used. However, in these studies, duplication detection was realized through the candidate video recorded on the server. This situation requires an ever-increasing storage and processing power requirement. The authors of \cite{uyan2022ti} improved the previous study and solved the mentioned storage and processing power problem by pulling the copy detection to the user side. Therefore, artificial intelligence tools used for copy detection work with the processor power of the user's computer, and only copy moments are reported. Our study is a continuation of these studies. 

Our main contributions include the following:
\begin{itemize}
    \item Making face detection more stable by updating face.api,
    \item Using the object detection model without training,
    \item Adding an audio detection model that detects background sounds.
\end{itemize}

The rest of this paper is organized as follows. In Section \ref{ProposedMethod}, details of the proposed deep learning method are explained. Experiments of the proposed models are presented in Section \ref{Experiments}. Finally, we evaluate the experiments' results and propose potential advancements for future work in Section \ref{Conclusion}.

\section{Proposed Method}
\label{ProposedMethod}

The copy detection system we created works as client-based. There were several advantages of setting up the whole structure as client-based. These benefits are listed below:
\begin{itemize}
\item Server costs are removed because all models run on the candidate's computer.
\item Server risks that may occur with many candidates taking the exam at the same time are prevented.
\item Since the candidates' data is only processed on their own computers, possible security breaches are prevented.
\item Thanks to the data from the candidate does not need to be sent to the server, machine-learning models can work instantly.
\end{itemize}

On the other hand, although the client-based system has many advantages since these models run on the candidate's computer, the working performance of these computers affects the working speed of the models. For this reason, we determined the image sampling rate to process a maximum of 3 frames per second. It could have increased the accuracy of our higher frame per second (FPS) models, but we kept the FPS quite low so that candidates who could enter with a low-performance computer would be acceptable during the exam. Despite this, we can get outstanding results in detecting duplicates. We will share these results in Section \ref{Experiments}. In addition, the image's resolution from the camera was determined as 400 pixels in width and 224 pixels in height. Keeping the resolution of the camera low was also for the same reasons we kept the FPS low.

Furthermore, we tried to choose models with tiny sizes in our system. We also run specific features in the pipeline when certain situations occur. We will describe these situations under Section \ref{ProposedMethod}.

\subsection{The Pipeline of Web-Client Based Cheating Detection System}

For machine-learning models to work, the candidate must keep his webcam and speaker on throughout the exam. After these permissions are given, specific issues are considered as suspected cheating. These behaviors are listed below:
\begin{itemize}
\item Using a mobile phone or laptop,
\item Leaving the exam environment
\item When someone else comes to the candidate during the exam,
\item Presence of someone else instead of the candidate in the exam,
\item Hearing the speaking voice during the exam.
  
\end{itemize}

\begin{figure*}[htp!]
	\centering
	\includegraphics[scale=0.105]{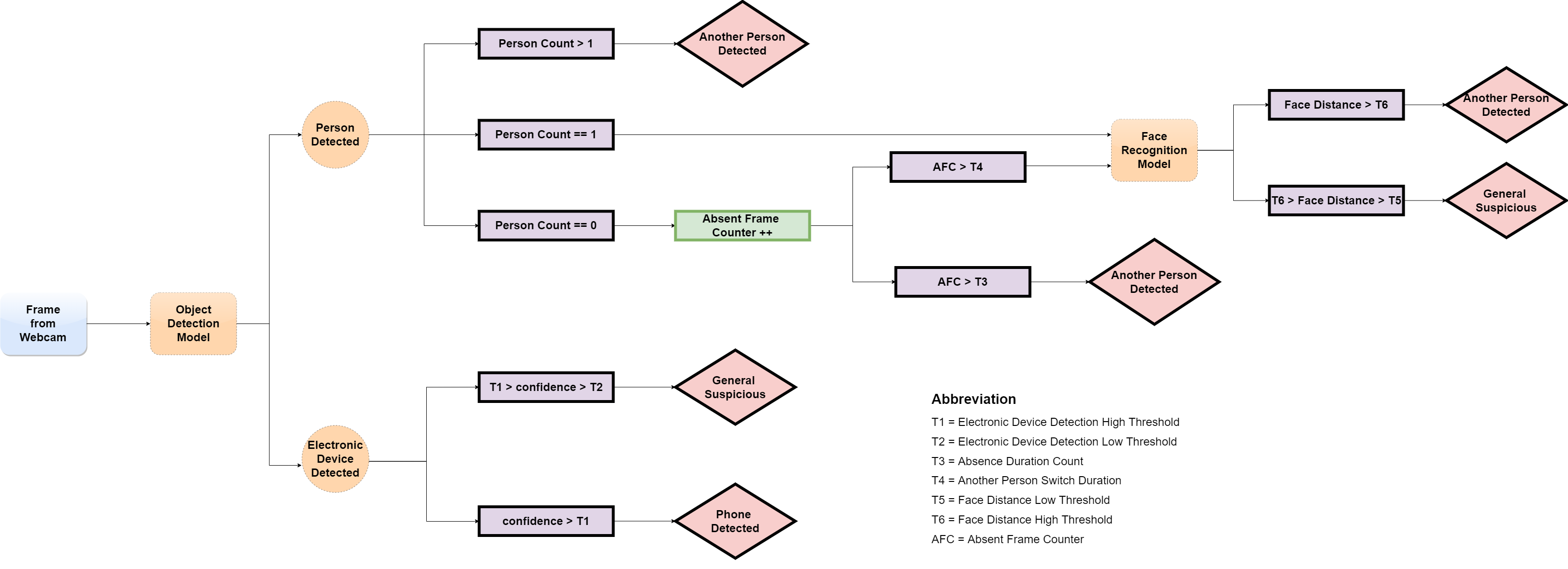}
	\caption{Pipeline of Cheating Detection System}
	\label{fig:PipelineOfTalentInterview}
\end{figure*}

If at least one of these items is fulfilled, a "Suspect" label is printed at the end of the exam. If none of these duplication conditions occur, the candidate's report is labeled "Clean". Furthermore, the cheating detection system pipeline is shown in Figure \ref{fig:PipelineOfTalentInterview}.

\subsection{Face Recognition}

Extracting facial features is an essential feature to detect possible duplicates. By creating the 128-dimensional characteristic of the face, it can be determined whether someone else took the exam instead of the candidate. Before the exam starts, the candidate's face is kept in 20 frames to extract the candidate's facial characteristics. 128-dimensional vectors are obtained from each of these. To obtain this vector, the "face-api.js" library was prepared in JavaScript and designed to extract its characteristics.

The 20 reference frames obtained before the exam were compared with the facial characteristics taken in the exam. In our client-based model, keeping the face characteristic for each frame during the exam will force the computer of the candidate taking the exam. Therefore, when the candidate is not seen on the screen and returns to the screen, the face characteristic is extracted and compared with the reference face characteristics. In this way, the necessary model was run in cases where someone else could take the exam instead of constantly extracting facial characteristics throughout the exam. The pipeline shows in which situations face recognition works can be observed in Figure \ref{fig:PipelineOfTalentInterview}.

While comparing the face with the reference frames, we looked at the Euclidean distances of the 128-dimensional vectors. We obtain a different Euclidean value for each reference. As soon as we observe that the lowest of these distances is higher than our threshold value, we present the "Another Person" cheating label in the report, which we determined as 0.6 as a result of experimental studies.

To explain labeling "Another Person" in detail, let $F_t$ be the face characteristic taken from a frame in the exam and $F_n$ is one of the reference vectors. We calculate this 128-dimensional vector with each reference vector that can be seen in Equation \ref{EuclideanDist}.

\begin{equation}
     d\left( F_{t},F_{n}\right)   = \sqrt {\sum _{i=1}^{128}  \left( F_{t_i}-F_{n_i}\right)^2 }.
     \label{EuclideanDist}
\end{equation}

Then, we find the minimum value in each Euclidean calculation, like the Equation \ref{anotherPerson}, and compare it with the threshold we set as 0.6.

\begin{equation}
X(\omega) = \begin{cases}
Clean & min(d) 	\leq 0.6 \\
Another Person & min(d) > 0.6,
\end{cases}
\label{anotherPerson}
\end{equation}
where
\begin{equation}
    min(d) = min(d\left( F_{t},F_{1}\right),..., d\left( F_{t},F_{n}\right), d\left( F_{t},F_{20}\right)).
    \label{minDist}
\end{equation}

\subsection{Object Detection}

Another type of cheating that may occur during the exam is using a phone or laptop. In the exam, the candidate can use the internet to find the answer to questions or share the answer with his friends. As a possible screen change can be tracked on the computer on which he/she takes the exam, the candidate may use electronic devices to cheat.

To prevent this, electronic objects used by the candidate are tried to be detected using the candidate's webcam. The COCO-SSD Lite MobilenetV2 \cite{sandler2018mobilenetv2} model in the TensorflowJS \cite{smilkov2019tensorflow} library designed as client-based was used. This model tries to classify 80 objects using the COCO \cite{lin2014microsoft} dataset. While performing feature extraction with MobilenetV2, objects are separated into bounding boxes using SSD \cite{liu2016ssd}. Since our intended use does not include a bounding box, if the camera sees a mobile phone or laptop and the electronic device detection score is greater than a certain threshold by the model, we write a "Suspect" label on the candidate's report.

We use two different thresholds to determine the electronic device's copy label. These are "phone detection low threshold" and "phone detection high threshold". If the probability of finding a mobile phone or laptop from the model is less than "phone detection low threshold", we will not detect it as a copy status. Still, if it is greater than "phone detection low threshold" and less than "phone detection high threshold", we will set it as "General Suspicious". On the other hand, we label it as "Phone Detection" if the probability of an electronic device from the model is higher than the "phone detection high threshold".

In addition to electronic devices, we use the COCO-SSD Lite MobilenetV2 model to detect person numbers in front of the camera. If the number of people coming from the model is one during the exam, we label the "Clean" on the candidate's report at the end of the exam. On the other hand, if a person cannot be determined in the exam and the candidate is not seen in front of the camera for 10 seconds, we write the "Candidate Absence Status" label on the report. Furthermore, suppose the MobilenetV2 model does not observe the candidate for less than 10 seconds but for more than 5 seconds. In that case, the candidate's 128-dimensional reference vectors generated at the beginning of the exam are compared with the vectors produced at that time, as in Equation \ref{anotherPerson}, and it is checked whether someone else has come to the exam.

\subsection{Human Voice Detection}

We tried to detect possible cheating operations that the webcam cannot reach, but that can be done with speech using the microphone. For this, we asked the candidate to open the microphone at the beginning of the exam. Afterward, we classify the sounds coming into the microphone as human or not human in the model we train beforehand with 1-second samples.

\begin{figure}[h]
	\centering
	\includegraphics[scale=0.125]{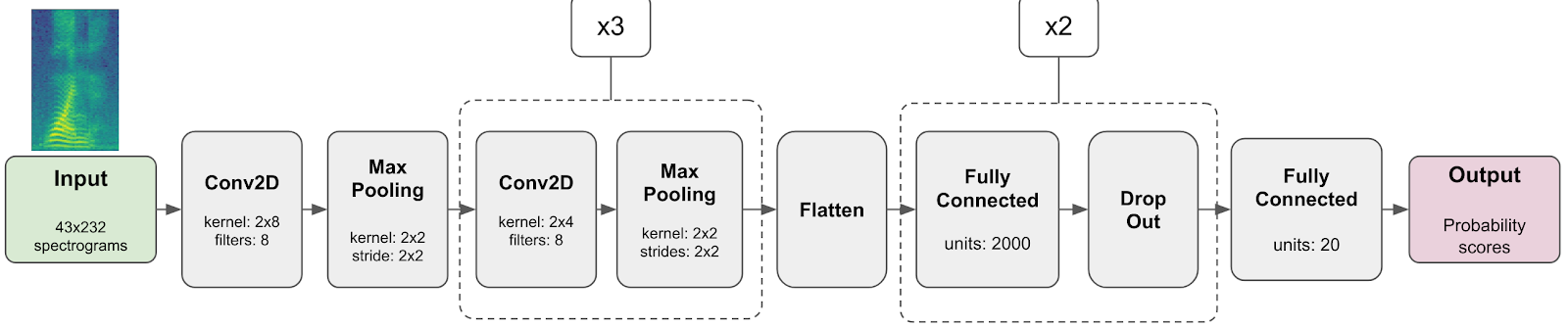}
	\caption{Dataflow of Human Sound Classification Architecture}
	\label{fig:soundArchitecture}
\end{figure}

While training this model, we used the human voice dataset \cite{chung2018voxceleb2} and the human, non-speech dataset \cite{piczak2015esc}. Before using the sounds in the datasets as input in the model, 1-second samples were taken and converted to Spectrogram. In this way, we obtained a two-dimensional input, and the Deep Learning architecture in Figure \ref{fig:soundArchitecture} was trained.

We deployed the model we obtained as client-based in JavaScript, using the TensorflowJS \cite{smilkov2019tensorflow} library. We used the browser's native Fourier Transform to convert the sounds from the candidate to Spectrogram during the exam. In this way, instead of listening and recording all the sounds, we labeled the sounds classified as "Voice Sound" in the Deep Learning model, where we only train the output of the "Browser FFT" as "Voice Detection".

\subsection{Segmentation}

During the models mentioned above are run, if a copy status is detected, a 5-second video is recorded from the moment it is detected. Since this video is taken from the webcam, the environment on the back of the candidate is also visible. This situation violates the General Data Protection Regulation (GDPR) and Personal Data Protection Authority rules. To eliminate this violation, we blur the places other than the candidate before recording the video.

We used the BodyPix MobileNet model in the TensorflowJS library to perform the specified operation. The MobileNet \cite{howard2017mobilenets} part performs feature extraction, while the BodyPix \cite{papandreou2018personlab} part applies the segmentation.

This model blurs all non-human things visible on the camera. However, if there is someone else with the candidate, we need to blur the people other than the candidate. To do this, we determine whether at least 4 of the 5 points (eyes, nose, two points of mouth) of the candidate's face from the face-api.js library are inside the face of the candidate from BodyPix. This way, we find the candidate's identity and blur the others.

\section{Experiments}
\label{Experiments}
Our proposed method for finding cheating, which can be seen in Figure \ref{fig:PipelineOfTalentInterview}, includes different sub-models.
Object classification and human voice classification are some of these methods.
Experiments have been made to measure the performance of these methods, and all experiments have been performed using Tesla P100 GPU or NVIDIA Tensor Core GPU machines in Google Colab Pro application.
\subsection{Human Voice Classification}

Before using the dataset \cite{chung2018voxceleb2} \cite{piczak2015esc} to be used to distinguish the human voice, three parts are created from this dataset: Training, validation, and testing. Training is done using the training data set. The validation data is selected from the training data and we use this dataset for the "Early-Stopping" method. When the validation loss value increases, we stop the training process with the "Early-Stopping" method. In this way, "overfitting" is prevented. After completing the training process, we use the test data to test our model.

We use the five-fold cross-validation method to test our model. Therefore, five equal parts are created from the data set. The first part is assigned as the test dataset, completely separated from the training and validation data to avoid data leakage. This process continues five times; another piece is assigned as test data each time. By taking the average of the results of this process, the problem of bias that may occur due to the fragmentation of the data set is prevented. In addition, to increase the reliability of the result, the five-fold cross-validation method was repeated three times, and the average of all results was taken.

\begin{table}[ph]
  \centering
  \caption{\textsc{Classification results of the human voice}}
  \label{tablo1}
  \begin{tabular}{|c|c|c|c|}
    \hline
    \multirow{2}{*}{} & \multicolumn{3}{c|}{Test Accuracy} \\
    \cline{2-4}
    & Min & Max & Mean \\
    \hline
     Human Voice Classification Architecture & 96.65 & 98.41 & 97.12 \\
    \hline
  \end{tabular}
\end{table}

\subsection{Object Detection}
Object detection aims to identify three situations: People in front of the camera, phones, and computers. The person's poses in different positions in front of the camera were successfully detected. Even in cases where only a single part of the person was visible, such as hands, feet, or head, the person could still be detected. However, a decrease in Intersection over Union (IoU) values was observed in these situations.

Different ways of holding the phone were tested in the phone detection part. Higher precision was observed when the phone was more clearly visible to the camera. However, a decrease in detection was observed when the person placed the phone against their ear, making it less visible. In some cases where only half of the phone was visible, it was mistaken for a remote control.

In the laptop detection part, the front of the laptop and cases where it had a more distinguishable shape were successfully detected. However, detection became more challenging when only a part of the laptop was visible. The front of the laptop is perceived with higher precision because it can extract more features.

\begin{table}[ph]
  \centering
  \caption{\textsc{Object Detection Accuracy Results}}
  \label{tablo2}
  \begin{tabular}{|c|c|c|c|}
    \hline
    \multirow{2}{*}{} & \multicolumn{3}{c|}{Test Accuracy} \\
    \cline{2-4}
     &Person & Phone & Laptop \\
    \hline
     Object Detection Model & 97.54 & 75.39 & 72.03 \\
    \hline
  \end{tabular}
\end{table}

For the tests, 1000 data points were used for each label, and a separate IoU threshold was assigned to each label. Based on this threshold, the model's accuracy on these labels was measured. The IoU thresholds used were 0.7 for people, 0.5 for laptops, and 0.3 for phones. The accuracy values obtained are 0.85 for person detection, 0.72 for laptop detection, and 0.63 for phone detection.

\section{Conclusion}
\label{Conclusion}
This article proposes a model that can run on a web client to avoid cheating during an online exam.
As far as we know, the system we recommend is the first to run on a web client and detect a copy.

As far as we know, it is also the only copy-detection system in which there is a classification of the human voice. The proposed human voice classification system has an average accuracy of 97.12. 

In addition, our object classification model, customized for copy detection, has a person-finding rate of 97.54, a phone-finding rate of 75.39, and a laptop-finding rate of 72.03. The system, in which all the sub-models are explained in detail, will be a good reference for studies on automatic copy detection.

\bibliographystyle{unsrt}  

\end{document}